\documentclass{article} 
\usepackage{nips13submit_e,times}
\usepackage{hyperref}
\usepackage{amsmath,graphics,graphicx}
\usepackage{cite}
\usepackage{url}

\title{Bidirectional Recursive Neural Networks\\ for Token-Level Labeling with Structure}

\author{
Ozan \.Irsoy \\
Department of Computer Science\\
Cornell University\\
Ithaca, NY 14853 \\
\texttt{oirsoy@cs.cornell.edu} \\
\And
Claire Cardie \\
Department of Computer Science\\
Cornell University \\
Ithaca, NY 14853 \\
\texttt{cardie@cs.cornell.edu} \\
}

\nipsfinalcopy 

\begin{document}

\maketitle

\begin{abstract}
Recently, deep architectures, such as recurrent and recursive neural networks have been
successfully applied to various natural language processing tasks. Inspired by
bidirectional recurrent neural networks which use representations that summarize
the past and future around an instance, we propose a novel architecture that
aims to capture the structural information around an input, and use it to label
instances. We apply our method to the
task of opinion expression extraction, where we employ the binary parse tree of a 
sentence as the structure, and word vector representations as the initial representation
of a single token. We conduct preliminary experiments
to investigate its performance and compare it to the sequential approach.

\end{abstract}

\section{Introduction}

Deep connectionist architectures involve many layers of nonlinear information processing. This 
allows them to incorporate meaning representations such that each succeeding layer potentially has a more
abstract meaning.
Recent
advancements on efficiently training deep neural networks enabled their application to many problems, including those in natural language processing (NLP). A key advance for application to NLP tasks was the invention of word embeddings that represent a single word as a 
dense, low-dimensional vector in a meaning space~\cite{bengioNeural}, from which numerous problems have
benefited~\cite{collobert-weston, collobertScratch}. In this work, we are interested in deep learning
approaches for NLP sequence tagging tasks in which the goal is to label each token of the given input sentence.  

Recurrent neural networks constitute one important class of naturally deep architecture that has been applied to many
sequential prediction tasks. In the context of NLP, recurrent neural networks view
a sentence as a sequence of tokens. With this view, they have been successfully applied to tasks such as language modeling~\cite{mikolovLanguage}, and spoken language understanding~\cite{slurnn}. Since  classical recurrent neural networks
only incorporate information from the past (i.e.\ preceding tokens),  bidirectional variants have been proposed to incorporate information from
both the past and the future (i.e.\ following tokens)~\cite{brecurrent}. Bidirectionality is especially useful for NLP tasks, since
information provided by the following tokens is usually helpful when making a decision on the current token.

Even though bidirectional recurrent neural networks rely on information from both preceding and following words, capturing long term dependencies might be difficult, due to the vanishing gradient problem~\cite{bengioVanishingGradient}: relevant information that is distant from the token under investigation might be lost.
On the other hand, depending on the task, a relevant token might be structurally close to the token under investigation, even though it is far away in sequence. As an example, a verb and its corresponding object might be
far away in terms of tokens if there are many adjectives before the object, but they would be
very close in the parse tree. 
In addition to a distance-based argument, structure also provides a different way of computing representations: it
allows for compositionality, i.e.\ the meaning of a phrase is determined via a composition of the meanings of the words that comprise it.
As a result, we believe that many NLP tasks might benefit from explicitly incorporating the structural information associated with a token. 

Recursive neural networks compose another class of architecture, one that operates on structured inputs. They have been applied to parsing 
~\cite{socher2011parsing}, sentence-level sentiment analysis ~\cite{socherSentiment}, and paraphrase detection ~\cite{socherUnfolding}. 
Given the structural representation of a sentence, e.g.\ a parse tree, they recursively generate
parent representations in a bottom-up fashion, by combining tokens to produce representations for phrases, eventually producing the whole sentence.  The sentence-level representation (or, alternatively, its phrases) is then used
to make a final classification for a given input sentence --- e.g.\ whether it conveys a positive or a negative sentiment. Since recursive neural networks generate
representations only for the internal nodes in the structured representation, they are not directly applicable to the token-level labeling tasks in which we are interested.

To this end,
we propose and explore an architecture that aims to represent the structural information associated with a single token.
In particular, we extend the traditional recursive neural network framework so that it not only generates representations for subtrees (i.e.\ phrases) and the whole sentence upward,
but also propagates downward representations toward the leaves,  carrying information about the structural environment of
each word.

Our method is naturally applicable to any type of labeling task at the word level, however we limit ourselves
to an opinion expression extraction task in this work. In addition, although the method is applicable to any type of positional directed
acyclic graph structure (e.g.\ the dependency parse of a sentence), we limit our attention to binary parse trees~\cite{socher2011parsing}. 

\section{Preliminaries}

\subsection{Task description}

\begin{table}[t]
\caption{An example sentence with labels}
\begin{center}
\begin{tabular}{|l|l|l|l|l|l|l|l|l|}
\hline
The & Committee &, &as &usual &, &and &as \\
\hline
B\_HOLDER & I\_HOLDER & O & B\_ESE & I\_ESE & O & O & O \\
\hline
\end{tabular}
\begin{tabular}{|l|l|l|l|l|l|l|l|l|l|}
\hline
expected &by &a &group &of &activists &and &blog &authors&,\\
\hline
O & O & O & O & O & O & O & O & O & O \\
\hline
\end{tabular}
\begin{tabular}{|l|l|l|l|l|l|l|}
\hline
has &refused &to &make &any &statements &.\\
\hline
B\_DSE & I\_DSE & I\_DSE & I\_DSE & I\_DSE & I\_DSE & O\\
\hline
\end{tabular}
\end{center}
\label{table:example}
\end{table}

Opinion expression identification aims to detect subjective expressions in a given text,
along with their characterizations, such as the intensity and sentiment of an opinion,
opinion holder and target, or topic ~\cite{wiebe2005}. This is important for tasks that require a fine-grained opinion
analysis, such as opinion oriented question answering and opinion summarization. In this work,
we focus on the detection of \emph{direct subjective expressions} (DSEs),
\emph{expressive subjective expressions} (ESEs) and opinion holders and targets, as defined in ~\cite{wiebe2005}. DSEs consist of
explicit mentions of private states or speech events expressing
private states, and ESEs consist of expressions that indicate sentiment, emotion, etc., without explicitly conveying them.
An example sentence with the appropriate labels is given in Table \ref{table:example} in which the DSE ``has refused to make any statements" explicitly expresses an opinion HOLDER's attitude and the ESE ``as usual" indirectly expresses the attitude of the writer.

Previously, opinion extraction has been tackled as a sequence labeling problem. This approach views a sentence as a sequence of tokens labeled using the conventional BIO tagging scheme: {\sc B} indicates the beginning of an opinion-related expression, {\sc I} is used for tokens inside the opinion expression, and {\sc O} indicates tokens outside any opinion-related class. Variants of conditional random field-based approaches have been successfully applied with this view~\cite{choi, breck, yang}.

Similar to CRF-based methods, recurrent neural networks can be applied to the problem of opinion
expression extraction, with a sequential interpretation. However this approach ignores the structural
information that is present in a sentence. Therefore we explore an architecture that incorporates structural information into the final decision.

\begin{figure}[t]
\includegraphics[width=\textwidth]{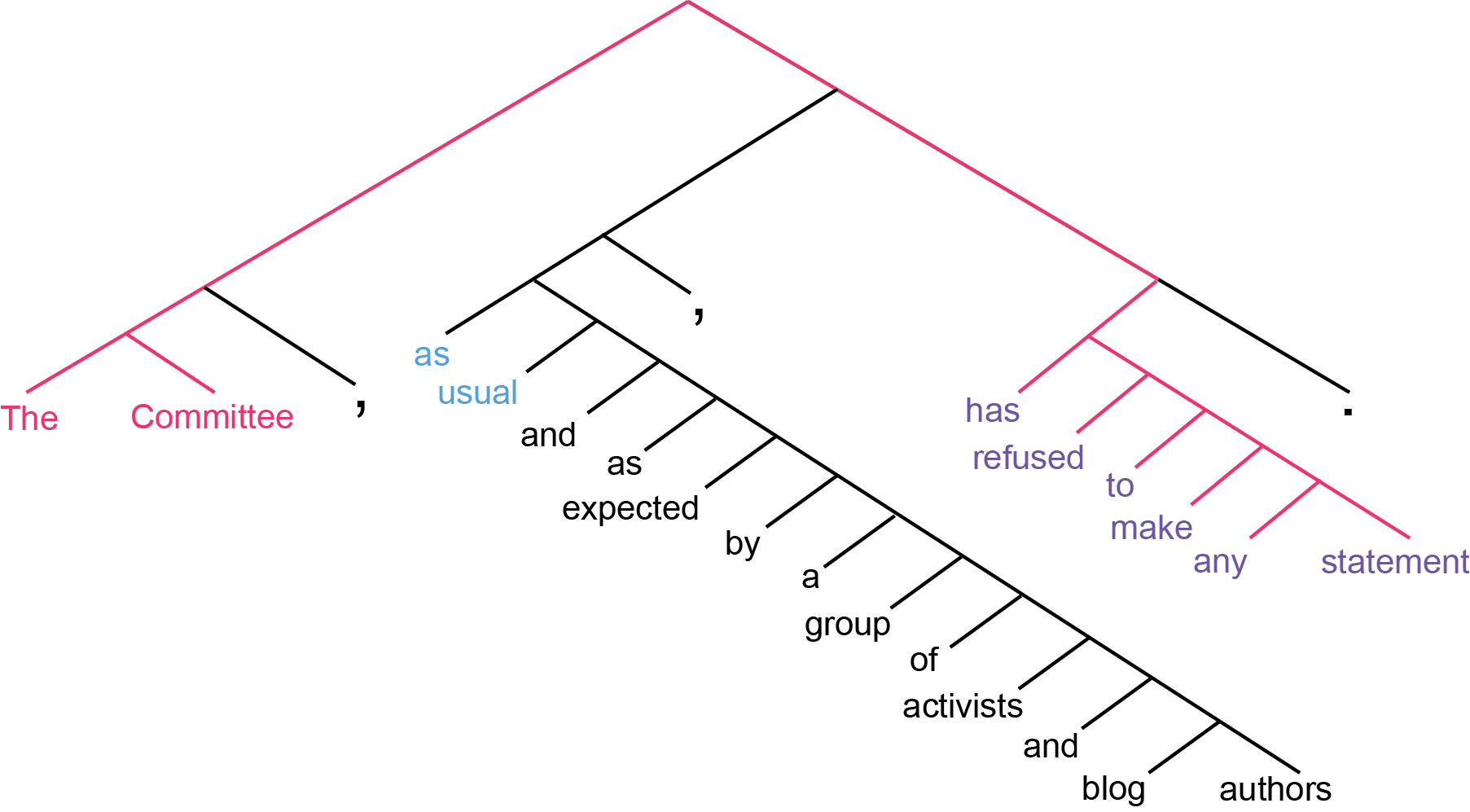}
\caption{Binary parse tree of the example sentence}
\label{fig:tree}
\end{figure}

\subsection{Word vector representations}

Neural network-based approaches require a vector representation for each input token. In natural language
processing, a common way of representing
a single token as a vector is to use a ``one-hot'' vector per token, with a dimensionality of the vocabulary
size, such that the corresponding entry of the vector is 1, and all others are 0. This results in a very
high dimensional, sparse representation. Alternatively, a distributed representation maps a token
to a real valued dense vector of smaller size (usually on the order of 100 dimensions). Generally,
these representations are learned in an unsupervised manner from a large corpus, e.g. Wikipedia. Various
architectures have been explored to learn these embeddings ~\cite{bengioNeural, collobert-weston, hlbl, mikolov}
which might have different generalization capabilities depending on the task ~\cite{turian}. The
geometry of the induced word vector space might have interesting semantic properties ~\cite{mikolov}.
In this work, we employ such word vector representations.

\subsection{Recurrent neural networks}
\label{sec:recurrent}

A recurrent neural network is a class of neural network that has recurrent connections, which allows a form of memory. This makes them applicable for sequential prediction tasks with arbitrary spatio-temporal dimension. This description
fits many natural language processing tasks, when a single
sentence is viewed as a sequence of tokens.
In this work,
we focus our attention on only the Elman type networks ~\cite{elman}.

In the Elman-type network, the hidden layer $h_t$ at time step $t$
is computed from a nonlinear transformation of the
current input layer
$x_t$ and the previous hidden layer $h_{t-1}$. Then, the
final output $y_t$ is computed using the hidden layer $h_t$.
One can interpret $h_t$ as an intermediate representation
summarizing the past, which
is used to make a final decision on the current input.
More formally,
\begin{align}
h_t &= f(W x_t + V h_{t-1} + b)\\
y_t &= g(W^o h_t + b^o)
\end{align}
where $f$ is a nonlinearity, such as the sigmoid function,
$g$ is the output nonlinearity, such as the softmax function,
$W$ and $V$ are weight matrices between the input and 
hidden layer, and among the hidden units themselves
(connecting the previous intermediate representation to the
current one), respectively, while $W^o$ is the output weight matrix, and
$b$ and $b^o$ are bias vectors connected to hidden and output units,
respectively.

Observe that with this definition, we have information only about
the past, when making a decision on $x_t$. This is limiting for most tasks.
A simple way to work around this problem is to include a fixed size
future context
around a single input vector. However this approach requires
tuning the context size, and ignores future information from outside of
the context window. Another way to incorporate information about the future
is to add bidirectionality to the architecture ~\cite{brecurrent}:
\begin{align}
h^\rightarrow_t &= f(W^\rightarrow x_t + V^\rightarrow h_{t-1} + b^\rightarrow)\\
h^\leftarrow_t &= f(W^\leftarrow x_t + V^\leftarrow h_{t+1} + b^\leftarrow)\\
y_t &= g(W^o_\rightarrow h^\rightarrow_t + W^o_\leftarrow h^\leftarrow_t + W^o x_t + b^o)
\end{align}
where $W^\rightarrow$ and $V^\rightarrow$ are the 
forward weight matrices as before,
$W^\leftarrow$ and $V^\leftarrow$ are the backward counterparts of them,
$W^o_\rightarrow$, $W^o_\leftarrow$, $W^o$ are the output matrices, and
$b^\rightarrow$, $b^\leftarrow$, $b^o$ are biases.
In this setting $h^\rightarrow_t$ and $h^\leftarrow_t$ can be interpreted as a
summary of the past, and the future, respectively, around the time step $t$. 
When we make a decision
on an input vector, we employ both the initial representation $x_t$ and the
two intermediate representations $h^\rightarrow_t$ and $h^\leftarrow_t$ of the
past and the future. Therefore in the bidirectional case, we have perfect
information about the sequence (ignoring the practical difficulties about
capturing long term dependencies, caused by vanishing gradients), whereas
the classical Elman type network uses only partial information.

Note that forward and backward parts of the network
are independent of each other until the output layer. This means that during training,
after backpropagating the error terms from the output layer to the hidden layer,
the two parts can be thought as separate, and trained with the classical
backpropagation through time~\cite{bptt}.

\subsection{Recursive neural networks}

Recursive neural networks comprise an architecture in which the same
set of weights is recursively applied in a structural setting: given a positional
directed acyclic graph, it visits the nodes in a topological order,
and recursively applies transformations to generate further representations
from previously computed representations of children. In fact, a recurrent neural
network is simply a recursive neural network with a particular structure.
Even though they can be applied to any positional directed acyclic graph, we limit
our attention to recursive neural networks over positional binary trees, 
as in~\cite{socher2011parsing}.

Given a binary tree structure with leaves having the initial representations,
e.g. a parse tree with word vector representations at the leaves, a recursive
neural network computes the representations at the internal node $\eta$ as follows:
\begin{align}
x_\eta = f(W_L x_{l(\eta)} + W_R x_{r(\eta)} + b)
\label{eqn:recursive}
\end{align}
where $l(\eta)$ and $r(\eta)$ are the left and right children of $\eta$,
$W_L$ an $W_R$ are the weight matrices that connect the left and right children
to the parent, and $b$ is a bias vector. Given that $W_L$ and $W_R$ are square
matrices, and not distinguishing whether $l(\eta)$ and $r(\eta)$ are leaf or
internal nodes, this definition has an interesting interpretation: initial
representations at the leaves and intermediate representation at the nonterminals
lie in the same space. In the parse tree example, recursive neural network 
combines representations of two subphrases to generate a representation for the larger phrase,
in the same meaning space ~\cite{socher2011parsing}. Depending on the task, we have a final output layer
at the root $\rho$:
\begin{align}
y = g(W^o x_\rho + b^o)
\end{align}
where $W^o$ is the output weight matrix and $b^o$ is the bias vector to the output layer.
In a supervised task, supervision occurs at this layer. Thus, during learning, initial
error is incurred on $y$, backpropagated from the root, towards leaves ~\cite{bpts}.

\section{Methodology}
\subsection{Bidirectional recursive neural networks}
\label{sec:recursive}

We will extend the aforementioned definition of recursive neural networks, so that
it propagates information about the rest of the tree, to every leaf node, through
structure. This will allow us to make decisions at the leaf nodes, with a summary
of the surrounding structure.

First, we modify the notation in equation (\ref{eqn:recursive}) so that it represents an upward layer
through the tree:
\begin{align}
x^\uparrow_\eta = f(W^\uparrow_L x^\uparrow_{l(\eta)} + W^\uparrow_R x^\uparrow_{r(\eta)} + b^\uparrow)
\end{align}
Note that $x^\uparrow_\eta$ is simply the initial representation $x_\eta$ if $\eta$ is a leaf, similar to equation 
(\ref{eqn:recursive}). Next, we add a downward layer on top of this upward layer:
\begin{align}
x^\downarrow_\eta = 
\begin{cases}
f(W^\downarrow_L x^\downarrow_{p(\eta)} + V^\downarrow x^\uparrow_\eta + b^\downarrow), & \text{if }\eta\text{ is a left child} \\
f(W^\downarrow_R x^\downarrow_{p(\eta)} + V^\downarrow x^\uparrow_\eta + b^\downarrow), & \text{if }\eta\text{ is a right child} \\
f(V^\downarrow x^\uparrow_\eta + b^\downarrow), & \text{if }\eta\text{ is root }(\eta = \rho)
\end{cases}
\end{align}
where $p(\eta)$ is the parent of $\eta$, 
$W^\downarrow_L$ and $W^\downarrow_R$ are the weight matrices that connect the downward representations of parent to
that of its
left and right children, respectively, $V^\downarrow$ is the weight matrix that connects the upward representation to the 
downward representation for any node, and $b^\downarrow$ is a bias vector at the downward layer. Intuitively, for any node,
$x^\uparrow_\eta$ contains information about the subtree rooted at $\eta$, and $x^\downarrow_\eta$ contains information about
the rest of the tree, since every node in the tree has a contributon to the computation of $x^\downarrow_\eta$. Therefore
$x^\uparrow_\eta$ and $x^\downarrow_\eta$ can be thought as complete summaries of the structure around $\eta$. At the leaves,
we use an output layer to make a final decision:
\begin{align}
y_\eta = g(W^o_\downarrow x^\downarrow_\eta + W^o_\uparrow x^\uparrow_\eta + b^o)
\end{align}
where $W^o_\downarrow$ and $W^o_\uparrow$ are the output weight matrices and $b^o$ is the output bias vector. In a supervised task,
supervision occurs at the output layer. Then, during training, error backpropagates upwards, through the downward layer, and then
downwards, through the upward layer, employing the backpropagation through structure method ~\cite{bpts}. If desired, backpropagated
errors can be used to update the initial representation $x$, which allows the possibility of fine tuning the word vector
representations, in our setting.

Note that this definition is structurally similar to the unfolding recursive autoencoder ~\cite{socherUnfolding}. However the goals of
the two architectures are different. Unfolding recursive autoencoder downward propagates representations as well. However, the
intention is to reconstruct the initial representations. On the other hand, we want the downward representations
$x^\downarrow$ to be as
different as possible than the upward representations $x^\uparrow$, since our aim is to capture the information about the
rest of the tree rather than the particular subtree under investigation. Thus, the unfolding recursive autoencoder does not use
$x^\uparrow$ when computing $x^\downarrow$ (except at the root), whereas bidirectional recursive neural network does.

\subsection{Incorporating sequential context}
\label{sec:combined}

Depending on the task, one might want to employ the sequential context around each input vector as well,
if the task has the sequential view in addition to structure. To this end,
we can combine the bidirectional recurrent neural network with the bidirectional recursive 
neural network. This allows to use both the
sequential information (past and future), and the structural information around a token to produce a final decision:
\begin{align}
y_\eta = g(W^o_\rightarrow h^\rightarrow_\eta + W^o_\leftarrow h^\leftarrow_\eta + 
            W^o_\downarrow x^\downarrow_\eta + W^o_\uparrow x^\uparrow_\eta + b^o)
\end{align}
This architecture can be seen as an extension to both the recurrent and the recursive neural network. During training,
after the error term backpropagates through the output layer, individual errors per each of the combined architectures can be handled separately, which allows us to use the previously noted training methods per architecture.

\section{Experiments}

We cast the problem of detecting DSEs and ESEs as two separate 3-class classification problems.
We also experiment with joint detection of DSEs, opinion holders, and opinion target --- as a 7-class classification problem
with one {\sc O}utside class and one {\sc B}eginning and {\sc I}nside class for DSEs, opinion holders and opinion targets. We compare the bidirectional recurrent
neural network as described in Section \ref{sec:recurrent} ({\sc Bi-Recurrent}), the bidirectional recursive network as described in Section
\ref{sec:recursive} ({\sc Bi-Recursive}), and the combined architecture as described in Section \ref{sec:combined} ({\sc Combined}).
We use the Stanford PCFG parser to extract binary parse trees of sentences~\cite{stanford}.

We use precision, recall and F-measure for performance evaluation. Since the boundaries of expressions are hard to define
even for human annotators~\cite{wiebe2005}, we use two soft notions of the measures: \emph{Binary Overlap} counts every overlapping
match between a predicted and true expression as correct~\cite{breck, yang}, and \emph{Proportional Overlap} imparts 
a partial correctness, proportional to the overlapping amount, to each match~\cite{johansson, yang}.

We use the manually annotated MPQA corpus~\cite{wiebe2005}, which 
has 14492 sentences in total. For DSE and ESE detection, we separate 4492 sentences as a test set, and run
10-fold cross validation. For joint detection of opinion holder, DSE and target, we have 9471 manually annotated sentences,
and we separate 2471 as a test set, and run 10-fold cross validation. A validation set is used to pick the best regularization parameter, simply a coefficient that penalizes the L2 norm.

We use standard stochastic gradient descent, updating weights after minibatches of 80 sentences. We run 200 epochs for training. Furthermore,
we fix the learning rate for every architecture, instead of tuning with cross validation, since initial experiments showed that
in this setting, every architecture successfully converges without any oscillatory behavior.

As initial representations of tokens, we use pre-trained Collobert-Weston embeddings ~\cite{collobert-weston}. 
Initial experiments with fine tuning
the word vector representations presented severe overfitting, hence, we keep the word vectors fixed in the experiments.

We employ the standard softmax activation for the output layer: $g(x) = e^{x_i} / \sum_j e^{x_j}$. For the hidden layers
we use the rectifier linear activation: $f(x) = \max\{0, x\}$. Experimentally, rectifier activation gives better performance,
faster convergence, and sparse representations. Note that in the recursive network, we apply the same transformation to both the leaf nodes and the
internal nodes, with the interpretation that they belong in the same meaning space. Employing the rectifier units at the upward layer
causes the upward representations at the internal nodes to be always nonnegative and sparse, whereas the initial representations
are dense, and might have negative values, which causes a conflict. To test the impact of this, we experimented with the sigmoid activation at the upward layer and the rectifier activation at the downward layer, which caused a degradation in performance. Therefore,
at a loss of interpretation, we use the rectifier activation at both layers in our experiments.

The number of hidden layers per architecture is chosen so that every architecture to be compared has the same number of hidden units connected to the output layer as well as the same
input dimensionality.

\subsection{Results}

Experimental results for DSE and ESE detection are given in Tables \ref{table:dse} and \ref{table:ese}. For the recurrent network,
the topology $(a,b,c)$ means that it has input dimensionality $a$, forward hidden layer dimensionality $b$,
and backward dimensionality $c$. For the recursive network, $(a,b)$ means that it has input
dimensionality and upward layer dimensionality $a$ and a
downward layer dimensionality $b$. For the combined network, $(a,b,c,d)$ means an input and upward layer
dimensionality $a$, downward layer dimensionality $b$ and forward and backward layer dimensionalities $c$ and $d$.
Asterisk indicates that the performance is statistically significantly better than others in the group, with respect
to a two sided paired t-test with $\alpha =0.05$.

\begin{table}[t]
\caption{Experimental results for DSE detection}
\begin{center}
\begin{tabular}{|ll| lll| lll|}
\hline 
Architecture & Topology &  \multicolumn{3}{c|}{Proportional} & \multicolumn{3}{c|}{Binary}\\
\hline 
 & & Prec. & Recall & F1 & Prec. & Recall & F1\\
 \hline
Bi-Recurrent & (50, 75, 75) & 56.59* & 56.60* & 56.60* & 58.84 & 62.23 & 60.49 \\
Bi-Recursive & (50, 150) & 53.93 & 55.05 & 54.48 & 58.21 & 62.29 & 60.23 \\
Combined & (50, 50, 50, 50) & 54.22 & 53.25 & 53.73 & 58.59 & 62.72 & 60.59 \\
\hline
\end{tabular}
\end{center}
\label{table:dse}
\end{table}

\begin{table}[t]
\caption{Experimental results for ESE detection}
\begin{center}
\begin{tabular}{|ll| lll| lll|}
\hline 
Architecture & Topology &  \multicolumn{3}{c|}{Proportional} & \multicolumn{3}{c|}{Binary}\\
\hline 
 & & Prec. & Recall & F1 & Prec. & Recall & F1\\
 \hline
Bi-Recurrent & (50, 75, 75) & 45.69 & 53.72 & 49.38 & 52.13 & 65.43 & 58.03 \\
Bi-Recursive & (50, 150) & 42.64 & 53.49 & 47.45 & 47.15 & 71.19* & 56.73 \\
Combined & (50, 50, 50, 50) & 46.16* & 53.33 & 49.49 & 51.95 & 67.49 & 58.71* \\
\hline
\end{tabular}
\end{center}
\label{table:ese}
\end{table}

\begin{table}[t]
\caption{Experimental results for joint holder+DSE+target detection}
\begin{center}
\begin{tabular}{|ll| ll|ll|ll|}
\hline 
Architecture & Topology &  \multicolumn{2}{c|}{DSE F1} & \multicolumn{2}{c|}{Holder F1} & \multicolumn{2}{c|}{Target F1}\\
\hline 
 & & Prop. & Bin. & Prop. & Bin. & Prop & Bin.\\
 \hline
Bi-Recurrent & (50, 75, 75)  & 49.73 & 54.49 & 48.19 & 51.36 & 39.32* & 50.53 \\
Combined & (50, 50, 50, 50) & 50.04 & 54.88 & 49.06* & 52.20* & 38.58 & 49.77\\
\hline
\end{tabular}
\end{center}
\label{table:joint}
\end{table}

We observe that the bidirectional recurrent neural network ({\sc Bi-Recurrent}) has better performance than both the bidirectional recursive ({\sc Bi-Recursive}) and the {\sc Combined} architectures
on the task of DSE detection, with respect to the proportional overlap metrics (56.60 F-measure, compared to
54.48, and 53.73). We do not observe a significant difference with respect to the
binary overalp metrics. This might be explained by the fact that DSEs tend to be shorter (often even a single word, such as ``criticized'' or
``agrees''). Furthermore, since DSEs exhibit explicit subjectivity, they do not neccessarily require a contextual investigation
around the phrase. Most of the time, a DSE can be detected just by looking at the particular phrase.

On the task of ESE detection, the {\sc Combined} network has significantly better binary F-measure compared to others (58.71 compared to
58.03 and 56.73).
Furthermore, the {\sc Combined} network has significantly better proportional precision than the two other architectures, at an insignificant loss in
proportional recall. In terms of binary measures, the {\sc Bi-Recursive} network has low precision and high recall, which might suggest a
complementary behavior for the two architectures. ESEs tend to be longer relative to DSEs, which might explain the results. 
Aditionally, unlike DSEs, ESEs more often require contextual information for their interpretation. For instance, in the given example in
Table~\ref{table:example}, it is not clear that
``as usual'' should be labeled as an ESE, unless one looks at the context presented in the sentence.

Experimental results for joint detection of opinion holder, DSE and target, are given in Table~\ref{table:joint}
(not to be compared with Table \ref{table:dse}, since the datasets are different). Here, the {\sc Combined} architecture
has insignificantly better performance in detecting DSEs (50.04 and 54.88 proportional and binary F-measures,
compared to 49.73 and 54.49), and significantly better performance in detecting opinion holders 
(49.06 and 52.20 proportional and binary F-measures, compared to 48.19 and 51.36), whereas the
{\sc Bi-Recurrent} network is better in detecting targets (39.32 and 50.53 proportional and binary F-measures, compared to 38.58 and 49.77).
Again, a possible explanation might be a better utilization of contextual information.
To decide whether a named entity is an opinion holder or not, one must link (or fail to link) the entity to an opinion expression.
Therefore, it is not possible to decide just by looking at the particular named entity.

\begin{figure}[t]
\centering
\includegraphics[width=0.49\textwidth]{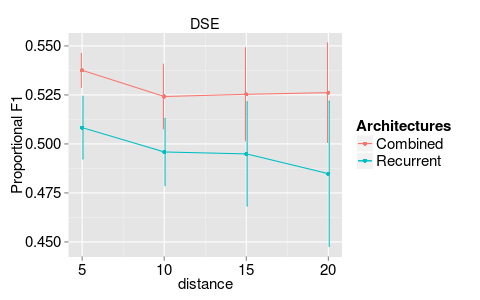}
\includegraphics[width=0.49\textwidth]{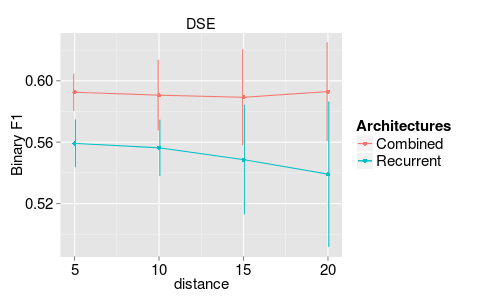}\\
\includegraphics[width=0.49\textwidth]{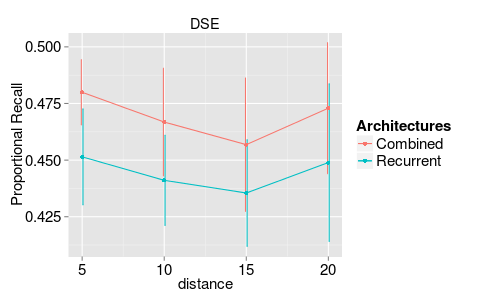}
\includegraphics[width=0.49\textwidth]{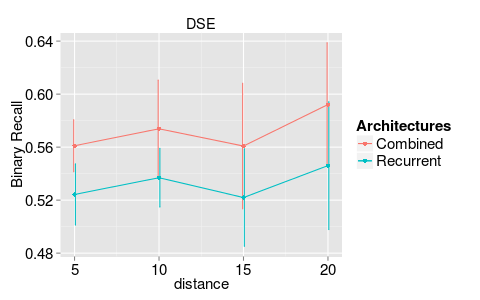}
\caption{Experimental results for joint detection over sentences with separation}
\label{fig:adt}
\end{figure}

For the joint detection task, we also investigate the performance on a subset of sentences, such that
each sentence has at least one DSE and opinion holder, and they are seperated by some distance. This is an attempt to
explore the impact of the token-level sequential distance between an opinion holder and an opinion expression. The results are given in
Figure~\ref{fig:adt}. As the separation distance increases, on average, DSE detection performance of the combined architecture is steady for the {\sc Combined} network compared
to the {\sc Bi-Recurrent} network. This might suggest that structural information helps to better capture the cues between opinion holders
and expression. Note that each distance-based subset of instances is strictly smaller, since there are fewer number of sentences conforming to the constraints, which causes an increase in variance.

\section{Conclusion and Discussion}

We have proposed an extension to the recursive neural network to carry out labeling tasks at the token level. We investigated its
performance on the opinion expression extraction task. Experiments showed that, depending on the task, employing the structural
information around a token might contribute to the performance.

In the bidirectional recursive neural network, downward layer is built on top of the upward layer, whereas in the bidirectional
recurrent neural network, forward and backward layers are separate. This causes the supervision to occur at a higher level
in the recursive network relative to the recurrent network, 
which makes training relatively more difficult. To alleviate this difficulty, an unsupervised
pre-training of the upward layer, or a similar semi-supervised training, as in ~\cite{socherSentiment}, 
might be employed as a future research direction.
A fine tuning of the word vector representations during this pre-training might have a positive impact on the performance
of the recursive network, since the learned representations for phrases might be structurally more meaningful,
compared to the representations learned by sequential, or context window based approaches. Future work will
address these observations,
investigate more effective training of the bidirectional recursive network and explore the impact of different word vector
representations on the architecture.

\subsubsection*{Acknowledgments}

This work was supported in part by DARPA DEFT Grant FA8750-13-2-0015
and a gift from Boeing.  The views and conclusions contained herein
are those of the authors and should not be interpreted as necessarily
representing the official policies or endorsements, either expressed
or implied, of DARPA or the U.S. Government. 

\bibliography{ref}{}
\small{
\bibliographystyle{unsrt}
}

\end{document}